\documentclass[10pt,twocolumn,letterpaper]{article}

\usepackage{cvpr}              

\usepackage{graphicx}
\usepackage{amsmath}
\usepackage{amssymb}
\usepackage{booktabs}

\usepackage{bbding}
\usepackage{multirow}
\usepackage{booktabs}
\usepackage[table,xcdraw,dvipsnames]{xcolor}
\usepackage{enumitem}

\usepackage[pagebackref,breaklinks,colorlinks]{hyperref}

\usepackage[capitalize]{cleveref}
\crefname{section}{Sec.}{Secs.}
\Crefname{section}{Section}{Sections}
\Crefname{table}{Table}{Tables}
\crefname{table}{Tab.}{Tabs.}

\def\cvprPaperID{2025} 
\def\confName{CVPR}
\def\confYear{2023}

\begin{document}

\title{Detecting and Grounding Multi-Modal Media Manipulation}


\author{
 Rui Shao$^{1,2}$\footnotemark[1], Tianxing Wu$^{2}$, Ziwei Liu$^{2}$\footnotemark[2]\\
 $^1$ School of Computer Science and Technology, Harbin Institute of Technology (Shenzhen) \\ 
 $^2$ S-Lab, Nanyang Technological University \\  
 {\tt\small shaorui@hit.edu.cn}, 
 {\tt\small\{tianxing.wu, ziwei.liu\}@ntu.edu.sg}\\
\texttt{\normalsize{\url{https://github.com/rshaojimmy/MultiModal-DeepFake}}}}

\twocolumn[{
\renewcommand\twocolumn[1][]{#1}

\maketitle
\begin{center}
\vspace*{-5mm}
  \centering
  \includegraphics[width=0.85\linewidth]{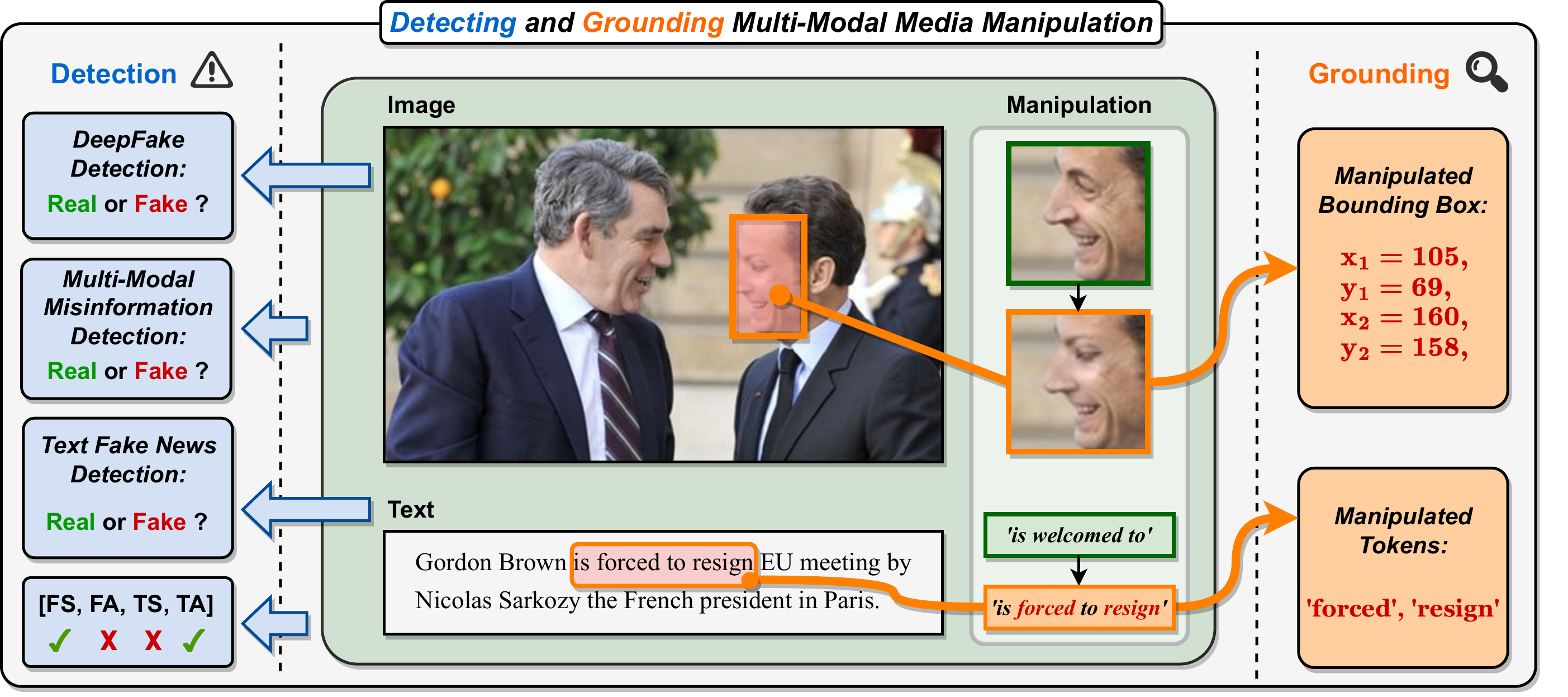}
  \vspace*{-2mm}
    \captionof{figure}{Different from existing single-modal forgery detection tasks, \textbf{DGM$^4$} not only performs real/fake classification on the input image-text pair, but also attempts to detect more fine-grained manipulation types and ground manipulated image bboxes and text tokens. They provide more comprehensive interpretation and deeper understanding about manipulation detection besides the binary classification. (FS: Face Swap Manipulation, FA: Face Attribute Manipulation, TS: Text Swap Manipulation, TA: Text Attribute Manipulation)}
  \label{fig:intro}
  \vspace*{-1mm}
\end{center}
}
]

\renewcommand{\thefootnote}{\fnsymbol{footnote}} 
\footnotetext[1]{This work was done at S-Lab, Nanyang Technological University}
\footnotetext[2]{Corresponding author}

\begin{abstract}
Misinformation has become a pressing issue. Fake media, in both visual and textual forms, is widespread on the web. 
While various deepfake detection and text fake news detection methods have been proposed, they are only designed for single-modality forgery based on binary classification, let alone analyzing and reasoning subtle forgery traces across different modalities. In this paper, we highlight a new research problem for multi-modal fake media, namely \textbf{D}etecting and \textbf{G}rounding \textbf{M}ulti-\textbf{M}odal \textbf{M}edia \textbf{M}anipulation (\textbf{DGM$^4$}). \textbf{DGM$^4$} aims to not only detect the authenticity of multi-modal media, but also ground the manipulated content (\textit{i.e.,} image bounding boxes and text tokens), which requires deeper reasoning of multi-modal media manipulation. To support a large-scale investigation, we construct the first \textbf{DGM$^4$} dataset, where image-text pairs are manipulated by various approaches, with rich annotation of diverse manipulations. Moreover, we propose a novel \textbf{H}ier\textbf{A}rchical \textbf{M}ulti-modal \textbf{M}anipulation r\textbf{E}asoning t\textbf{R}ansformer (\textbf{HAMMER}) to fully capture the fine-grained interaction between different modalities. \textbf{HAMMER} performs \textbf{1)} manipulation-aware contrastive learning between two uni-modal encoders as shallow manipulation reasoning, and \textbf{2)} modality-aware cross-attention by multi-modal aggregator as deep manipulation reasoning. Dedicated manipulation detection and grounding heads are integrated from shallow to deep levels based on the interacted multi-modal information. Finally, we build an extensive benchmark and set up rigorous evaluation metrics for this new research problem. Comprehensive experiments demonstrate the superiority of our model; several valuable observations are also revealed to facilitate future research in multi-modal media manipulation.
\end{abstract}
\vspace*{-7mm}
\section{Introduction}
\label{sec:intro}


\begin{table*}[t]
\scriptsize
\centering
\caption{Comparison of the proposed \textbf{DGM$^4$} with existing tasks related to image and text forgery detection.}
\vspace*{-3mm}
\begin{tabular}{lccccc}
\toprule[1pt]
\multicolumn{1}{c}{\multirow{2}{*}{\textbf{Problem Setting}}} & \multicolumn{2}{c}{\textbf{Image Forgery}}                    & \multicolumn{2}{c}{\textbf{Text Forgery}}                     & \multirow{2}{*}{\textbf{\begin{tabular}[c]{@{}c@{}}Multi-Modal\\ Forgery Detection\end{tabular}}} \\ \cline{2-5}
\multicolumn{1}{c}{}                                          & \textbf{Detection}            & \textbf{Grounding}            & \textbf{Detection}            & \textbf{Grounding}            &                                                                                                   \\ \hline
DeepFake Detection~\cite{luo2021generalizing,zhao2021multi}                                            & \CheckmarkBold &  \XSolidBrush  &  \XSolidBrush  &  \XSolidBrush  &     \XSolidBrush                                                                   \\
Text Fake News Detection~\cite{wang2017liar,zellers2019defending}                                     & \XSolidBrush   &  \XSolidBrush  & \CheckmarkBold &   \XSolidBrush &  \XSolidBrush                                                                      \\
Multi-Modal Misinformation Detection~\cite{abdelnabi2022open,luo2021newsclippings}                         & \XSolidBrush   & \XSolidBrush   & \XSolidBrush   & \XSolidBrush   & \CheckmarkBold                                                                     \\
\rowcolor[HTML]{E3DCDC} \textbf{DGM$^4$}        & \CheckmarkBold & \CheckmarkBold & \CheckmarkBold & \CheckmarkBold & \CheckmarkBold                                                                     \\ \bottomrule[1pt]
\end{tabular}
\label{tab:intro}
\vspace*{-4mm}
\end{table*}

With recent advances in deep generative models, increasing hyper-realistic face images or videos can be automatically generated, which results in various security issues~\cite{2018TIFSdynamictext,shao2022open,Shao_2019_CVPR,Shao_2020_AAAI,shao2020open,shao2022federated,shao2017deep,shao2021federated,yu2020fas} such as serious \textit{deepfake} problem~\cite{rossler2019faceforensics++,li2019celeb,dolhansky2020deepfake,jiang2020deeperforensics,shao2022detecting} spreading massive fabrication on visual media. This threat draws great attention in computer vision community and various deepfake detection methods have been proposed. With the advent of Large Language Model, \textit{e.g.}, BERT~\cite{devlin2019bert}, GPT~\cite{radford2019language}, enormous \textit{text fake news}~\cite{wang2017liar,zellers2019defending} can be readily generated to maliciously broadcast misleading information on textual media. Natural Language Processing (NLP) field pays great attention to this issue and presents diverse text fake news detection methods.

Compared to a single modality, the multi-modal media (in form of image-text pairs) disseminates broader information with greater impact in our daily life. Thus, multi-modal forgery media tends to be more harmful. To cope with this new threat with a more explainable and interpretable solution, this paper proposes a novel research problem, namely \textbf{D}etecting and \textbf{G}rounding \textbf{M}ulti-\textbf{M}odal \textbf{M}edia \textbf{M}anipulation (\textbf{DGM$^4$}). As shown in Table~\ref{tab:intro} and Fig.~\ref{fig:intro}, two challenges are brought by \textbf{DGM$^4$}: \textbf{1)} while current deepfake detection and text fake news detection methods are designed to detect forgeries of single modality, \textbf{DGM$^4$} demands simultaneously detecting the existence of forgery in both image and text modality and \textbf{2)} apart from binary classification like current single-modal forgery detection, \textbf{DGM$^4$} further takes grounding manipulated image bounding boxes (bboxes) and text tokens into account. This means existing single-modal methods are unavailable for this novel research problem. A more comprehensive and deeper reasoning of the manipulation characteristics between two modalities is of necessity. Note that some multi-modal misinformation works~\cite{abdelnabi2022open,luo2021newsclippings} are developed. But they are only required to determine binary classes of multi-modal media, let alone manipulation grounding.

To facilitate the study of \textbf{DGM$^4$}, this paper contributes the first large-scale \textbf{DGM$^4$} dataset. In this dataset, we study a representative multi-modal media form, \textit{human-centric news}. It usually involves misinformation regarding politicians and celebrities, resulting in serious negative influence. We develop two different image manipulation (\textit{i.e.,} face swap/attribute manipulation) and two text manipulation (\textit{i.e.,} text swap/attribute manipulation) approaches to form the multi-modal media manipulation scenario. Rich annotations are provided for detection and grounding, including binary labels, fine-grained manipulation types, manipulated image bboxes and manipulated text tokens. 

Compared to pristine image-text pairs, manipulated multi-modal media is bound to leave manipulation traces in manipulated image regions and text tokens. All of these traces together alter the cross-modal correlation and thus cause semantic inconsistency between two modalities. Therefore, reasoning semantic correlation between images and texts provides hints for the detection and grounding of multi-modal manipulation. To this end, inspired by existing vision-language representation learning works~\cite{li2021align,radford2021learning,kim2021vilt}, we propose a novel \textbf{H}ier\textbf{A}rchical \textbf{M}ulti-modal \textbf{M}anipulation r\textbf{E}asoning t\textbf{R}ansformer (\textbf{HAMMER}) to tackle \textbf{DGM$^4$}. To fully capture the interaction between images and texts, \textbf{HAMMER} \textbf{1)} aligns image and text embeddings through manipulation-aware contrastive learning between two uni-modal encoders as shallow manipulation reasoning and \textbf{2)} aggregates multi-modal embeddings via modality-aware cross-attention of multi-modal aggregator as deep manipulation reasoning. Based on the interacted multi-modal embeddings in different levels, dedicated manipulation detection and grounding heads are integrated hierarchically to detect binary classes, fine-grained manipulation types, and ground manipulated image bboxes, manipulated text tokens. This hierarchical mechanism contributes to more fine-grained and comprehensive manipulation detection and grounding. Main contributions of our paper:
\vspace*{-2mm}
\begin{itemize}[leftmargin=*]
\item We introduce a new research problem \textbf{D}etecting and \textbf{G}rounding \textbf{M}ulti-\textbf{M}odal \textbf{M}edia \textbf{M}anipulation (\textbf{DGM$^4$}), with the objective of detecting and grounding manipulations in image-text pairs of human-centric news.
\vspace*{-2mm}
\item We contribute a large-scale \textbf{DGM$^4$} dataset with samples generated by two image manipulation and two text manipulation approaches. Rich annotations are provided for detecting and grounding diverse manipulations.
\vspace*{-2mm}
\item We propose a powerful \textbf{H}ier\textbf{A}rchical \textbf{M}ulti-modal \textbf{M}anipulation r\textbf{E}asoning t\textbf{R}ansformer (\textbf{HAMMER}). A comprehensive benchmark is built based on rigorous evaluation protocols and metrics. Extensive quantitative and qualitative experiments demonstrate its superiority.
\end{itemize}

\section{Related Work}
\label{sec:related work}

\noindent \textbf{DeepFake Detection.}
To detect face forgery images, current deepfake detection methods are built based on spatial and frequency domains. Spatial-based deepfake detection methods exploit spatial visual cues, such as blending artifacts~\cite{li2020face}, textural features~\cite{chen2021local,zhao2021multi,zhu2021face}, 3D information~\cite{zhu2021face}, patch consistency~\cite{zhao2021learning} and noise characteristics~\cite{gu2022exploiting}. Frequency-based deepfake detection methods detect spectrum artifacts, like high-frequency components decomposed from Discrete Fourier Transform (DFT)~\cite{dzanic2020fourier}, subtle frequency discrepancy derived from Discrete Cosine Transform (DCT)~\cite{qian2020thinking}, up-sampling artifacts hidden in phase spectrum~\cite{liu2021spatial} and frequency-based metric learning~\cite{li2021frequency}. Most of the above deepfake detection methods only perform binary classification in image media, not to mention manipulation grounding across multi-modalities.


\noindent \textbf{Multi-Modal Misinformation Detection.}
Several existing works study the detection of multi-modal misinformation~\cite{khattar2019mvae,jin2017multimodal,aneja2021cosmos,luo2021newsclippings,wang2018eann,abdelnabi2022open}. Some of them deal with a small-scale human-generated multi-modal fake news~\cite{khattar2019mvae,jin2017multimodal,wang2018eann}, while others address out-of-context misinformation where a real image is paired with another swapped text without image and text manipulation~\cite{aneja2021cosmos,luo2021newsclippings,abdelnabi2022open}. All of these methods only perform binary classification based on simple image-text correlation. In contrast, \textbf{DGM$^4$} studies large-scale machine-generated multi-media manipulation, which is closer to broad misinformation on the web in practice. Additionally, \textbf{DGM$^4$} requires not only manipulation detection for binary classification, but also manipulation grounding with more interpretation for multi-modal manipulation.


\section{Multi-Modal Media Manipulation Dataset}
\label{sec:dataset}


Most of existing misinformation datasets focus on single-modal image forgery ~\cite{rossler2019faceforensics++,li2020celeb,dolhansky2020deepfake,jiang2020deeperforensics} or text forgery ~\cite{wang2017liar,shu2017fake,zellers2019defending}. Some multi-modal datasets are built, but they usually contain a small amount of human-generated fake news~\cite{boididou2018detection,jin2017multimodal} or  out-of-context pairs~\cite{aneja2021cosmos,luo2021newsclippings} for binary forgery detection. To better facilitate the proposed novel research problem, we present \textbf{DGM$^4$} dataset, studying large-scale machine-generated multi-modal media manipulation. \textbf{DGM$^4$} dataset is constructed with diverse manipulation techniques on both \textbf{Image} and \textbf{Text} modality. All samples are annotated with rich, fine-grained labels that enable both \textbf{Detection} and \textbf{Grounding} of media manipulation. 

\subsection{Source Data Collection}

Among all forms of multi-modal media, we specifically focus on \textit{human-centric news}, in consideration of its great public influence.
We thus develop our dataset based on the VisualNews dataset~\cite{liu2021visual}, which collects numerous image-text pairs from real-world news sources (The Guardian, BBC, USA TODAY, and The Washington Post). To formulate a human-centric scenario with meaningful context, we further conduct data filtering on both image and text modality, and only keep the appropriate pairs to form the source pool $O=\{p_o|p_o=(I_o, T_o)\}$ for manipulation. 
\subsection{Multi-Modal Media Manipulation}
\label{sec:Dataset_Manipulation}
\begin{figure*}[t] 
	\begin{center}
		\includegraphics[width=0.95\linewidth]{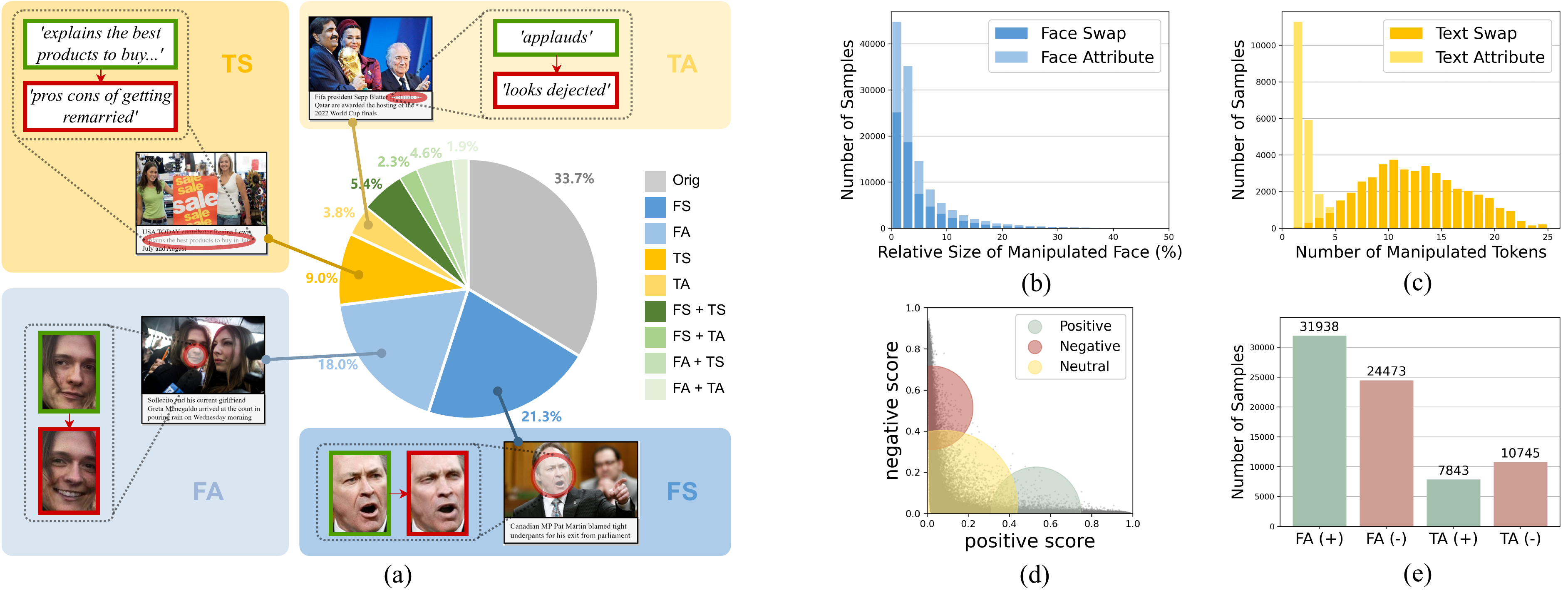}
	\end{center}
 	\vspace*{-7mm}
	\caption{Statistics of \textbf{DGM$^4$} dataset. (a) Distribution of manipulation classes; (b) manipulated regions of most images are small-size, especially for face attribute manipulation; (c) manipulated tokens of text attribute manipulation are fewer than text swap manipulation; (d) distribution of text sentiment scores in the source pool. (e) number of manipulated samples towards each face/text attribute direction.}
	\label{fig:statistics}
	\vspace*{-5mm}
\end{figure*}

We employ two types of harmful manipulations on both image and text modality. `Swap' type is designed to include relatively global manipulation traces, while `Attribute' type introduces more fine-grained local manipulations. The manipulated images and texts are then randomly mixed with pristine samples to form a total of 8 fake plus one original manipulation classes. Distribution of manipulation classes and some samples are displayed in Fig.~\ref{fig:statistics} (a).

\noindent \textbf{Face Swap (FS) Manipulation}. In this manipulation type, the \textit{identity} of the main character is attacked by swapping his/her face with another person. We adopt two representative face swap approaches, SimSwap~\cite{chen2020simswap} and InfoSwap~\cite{gao2021information}. For each original image $I_o$, we choose one of the two approach to swap the largest face $I_o^f$ with a random source face $I_{celeb}^f$ from CelebA-HQ dataset~\cite{karras2018progressive}, producing a face swap manipulation sample $I_{s}$. The MTCNN bbox of the swapped face $y_{\rm box}=\{ x_1, y_1, x_2, y_2\}$ is then saved as annotation for grounding. 

\noindent \textbf{Face Attribute (FA) Manipulation}. As a more fine-grained image manipulation scenario, face attribute manipulation attempts to manipulate the \textit{emotion} of the main character's face while preserving the identity. For example, if the original face is smiling, we deliberately edit it to the opposite emotion, \textit{e.g.,} an angry face. To achieve this, we first predict the original facial expression of the aligned face $I_o^f$ with a CNN-based network, then edit the face towards the opposite emotion using GAN-based methods, HFGI~\cite{wang2021HFGI} and StyleCLIP~\cite{patashnik2021styleclip}. After obtaining the manipulated face $I_{emo}^f$, we re-render it back to the original image $I_o$ to obtain the manipulated sample $I_{a}$. Bbox $y_{\rm box}$ is also provided.

\noindent \textbf{Text Swap (TS) Manipulation}. In this scenario, the text is manipulated by altering its overall \textit{semantic} while preserving words regarding main character. Given an original caption $T_o$, we use Named Entity Recognition (NER) model to extract the person's name as query `PER'. Then we retrieve a different text sample $T_o'$ containing same `PER' entity from the source corpus $O$. $T_o'$ is then selected as the manipulated text $T_s$. Note that we compute the semantic embedding of each text using Sentence-BERT~\cite{reimers2019sentence} and only accept $T_o'$ that has low cosine similarity with $T_o$. This ensures the retrieved text is not semantically aligned with $T_o$, so that the text semantic regarding the main character in the obtained pair $p_m=(I_o, T_s)$ is manipulated. After that, given $M$ text tokens in $T_s$, we annotate them with a $M$-dimensional one-hot vector $y_{\rm tok} = \{y_i\}^M_{i=1}$, where $y_i \in \{0, 1 \}$ denotes whether the $i$-th token in $T_s$ is manipulated or not.

\noindent \textbf{Text Attribute (TA) Manipulation}. Although news is a relatively objective media form, we observe that a considerable portion of news samples $p_o \in O$ still carry \textit{sentiment} bias within the text $T_o$, as depicted in Fig.~\ref{fig:statistics} (d). The malicious manipulation of text attributes, especially its sentiment tendency, could be more harmful and also harder to be detected as it causes less cross-modal inconsistency than text swap manipulation. To reflect this specific situation, we first use a RoBERTa~\cite{liu2019roberta} model to split the captions into positive, negative and neutral sentiment corpora: $\{O_{+}, O_{-}, O_{neu}\}$. Following \cite{Sudhakar2019TransformingDR}, we replace all sentiment words of the original text $T_o$ with the opposite sentiment text generated by a B-GST model trained on our own corpora $\{O_{+}, O_{-}\}$, obtaining $T_a$. Similar to text swap manipulation, all text tokens are also annotated with ground-truth vector $y_{\rm tok}$.

\noindent \textbf{Combination and Perturbation}. Once all single-modal manipulations are finished, we combine the obtained manipulation samples $I_s$, $I_a$, $T_s$ and $T_a$ with the original $(I_o, T_o)$ pairs. This forms a multi-modal manipulated media pool with full manipulation types: $P=\{p_m|p_m=(I_x, T_y), x,y\in\{o, s, a\}\}$. Each pair $p_m$ in the pool is provided with a binary label $y_{\rm bin}$, a fine-grained manipulation type annotation $y_{\rm mul}$, aforementioned annotations $y_{\rm box}$ and $y_{\rm tok}$. $y_{\rm bin}$ describes whether the image-text pair $p_m$ is real or fake, and $y_{\rm mul} = \{y_j\}^4_{j=1}$ is a 4-dimensional vector denoting whether the $j$-th manipulation type (\textit{i.e.,} FS, FA, TS, TA) appears in $p_m$.
To better reflect the real-world situation where manipulation traces may be covered up by noise, we employ random image perturbations on 50\% of the media pool $P$, such as JPEG compression, Gaussian Blur, \textit{etc}.


\begin{figure*}[t] 
	\begin{center}
		\includegraphics[width=1\linewidth]{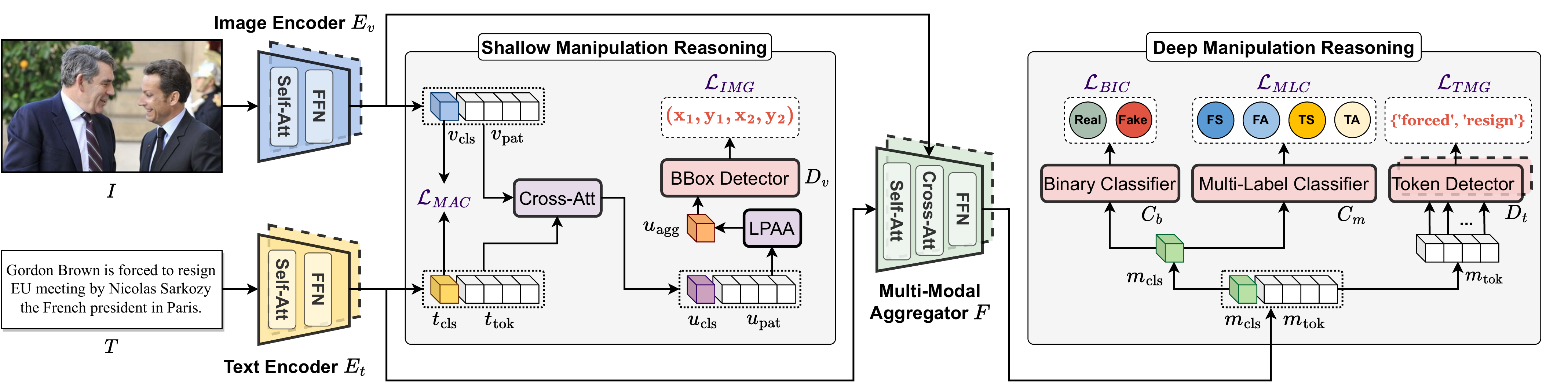}
	\end{center}
	\vspace*{-4mm}
	\caption{Overview of proposed \textbf{HAMMER}. It \textbf{1)} aligns image and text embeddings through manipulation-aware contrastive learning between Image Encoder $E_v$, Text Encoder $E_t$ in shallow manipulation reasoning and \textbf{2)} further aggregates multi-modal embeddings via modality-aware cross-attention of Multi-Modal Aggregator $F$ in deep manipulation reasoning. Based on the interacted multi-modal embeddings in different levels, various manipulation detection and grounding heads (Multi-Label Classifier $C_m$, Binary Classifier $C_b$, BBox Detector $D_v$, and Token Detector $D_t$) are integrated to perform their tasks hierarchically. Modules with dashed lines mean they are the corresponding momentum versions of Image Encoder, Text Encoder, Multi-Modal Aggregator and Token Detector, respectively.}
	\label{fig:framework}
	\vspace*{-4mm}
\end{figure*}

\subsection{Dataset Statistics}
\label{sec:Dataset_Statistics}
The overall statistics of \textbf{DGM$^4$} dataset are illustrated in Fig.~\ref{fig:statistics} (a). It consists a total of \textbf{230k} news samples, including 77,426 pristine image-text pairs and 152,574 manipulated pairs. The manipulated pairs contain 66,722 face swap manipulations, 56,411 face attribute manipulations, 43,546 text swap manipulations and 18,588 text attribute manipulations. $\sim$1/3 of the manipulated images and $\sim$1/2 of the manipulated text are combined together to form 32,693 mixed-manipulation pairs. Since both image and text attributes can be edited towards two opposite sentiment directions, we deliberately keep a balanced proportion to create an emotionally-balanced dataset, as shown in Fig.~\ref{fig:statistics} (e).

Furthermore, it can be observed from Fig.~\ref{fig:statistics} (b)-(c) that the manipulated regions of most images and the number of manipulated text tokens are relatively small. This indicates \textbf{DGM$^4$} dataset provides a much more challenging scenario for forgery detection compared to existing deepfake and multi-modal misinformation datasets. 

\section{HAMMER}
\label{sec:method}

To address \textbf{DGM$^4$}, as illustrated in Fig.~\ref{fig:framework}, we propose a \textbf{H}ier\textbf{A}rchical \textbf{M}ulti-modal \textbf{M}anipulation r\textbf{E}asoning t\textbf{R}ansformer (\textbf{HAMMER}), which is composed of two uni-modal encoders (\textit{i.e.,} Image Encoder $E_v$, Text Encoder $E_t$), Multi-Modal Aggregator $F$, and dedicated manipulation detection and grounding heads (\textit{i.e.,}  Binary Classifier $C_b$, Multi-Label Classifier $C_m$, BBox Detector $D_v$, and Token Detector $D_t$). All of these uni-modal encoders and multi-modal aggregator are built based on transformer-based architecture~\cite{vaswani2017attention}. As mentioned above, modeling semantic correlation and capturing semantic inconsistency between two modalities can facilitate detection and grounding of multi-modal manipulation. However, there exist two challenges \textbf{1)} as discussed in Sec.~\ref{sec:Dataset_Statistics} and shown in Fig.~\ref{fig:statistics} (b)-(c), a large portion of multi-modal manipulations are minor and subtle, locating in some small-size faces and a few word tokens and \textbf{2)} much visual and textual noise~\cite{li2021align} exists in multi-modal media on the web. As a result, some semantic inconsistencies caused by manipulation may be neglected or covered by noise. This demands more fine-grained reasoning of multi-modal correlation. To this end, we devise \textbf{HAMMER} to perform hierarchical manipulation reasoning which explores multi-modal interaction from shallow to deep levels, along with hierarchical manipulation detection and grounding. In the shallow manipulation reasoning, we carry out semantic alignment between image and text embeddings through Manipulation-Aware Contrastive Loss $\mathcal{L}_{MAC}$, and conduct manipulated bbox grounding under Image Manipulation Grounding Loss $\mathcal{L}_{IMG}$. In the deep manipulation reasoning, based on deeper interacted multi-modal information generated by Multi-Modal Aggregator, we then detect binary classes with Binary Classification Loss $\mathcal{L}_{BIC}$, fine-grained manipulation types with Multi-Label Classification Loss $\mathcal{L}_{MLC}$, and ground manipulated text tokens via Text Manipulation Grounding Loss $\mathcal{L}_{TMG}$. By combing all the above losses, manipulation reasoning is performed hierarchically, contributing to a joint optimization framework as follows:
\begin{equation}
\footnotesize
\begin{split}
\mathcal{L} = \mathcal{L}_{MAC} + \mathcal{L}_{IMG} + \mathcal{L}_{MLC} + \mathcal{L}_{BIC} + \mathcal{L}_{TMG}
\end{split}
\label{equ:opt} 
\end{equation}


\subsection{Shallow Manipulation Reasoning}

Given an image-text pair $(I, T) \sim P$, we patchify and encode image $I$ into a sequence of image embeddings via self-attention layers and feed-forward networks in Image Encoder as $E_v(I)=\{v_{\rm cls}, v_{\rm pat} \}$, where $v_{\rm cls}$ is the embedding of \texttt{[CLS]} token, and $v_{\rm pat} = \{v_1, ..., v_{\rm N}\}$ are embeddings of $N$ corresponding image patches. Text Encoder extracts a sequence of text embeddings of $T$ as $E_t(T)=\{t_{\rm cls}, t_{\rm tok} \}$, where $t_{\rm cls}$ is the embedding of \texttt{[CLS]} token, and $t_{\rm tok} = \{t_1, ..., t_{\rm M}\}$ are embeddings of $M$ text tokens. 

\noindent \textbf{Manipulation-Aware Contrastive Learning.}
To help two uni-modal encoders better exploit the semantic correlation of images and texts, we align image and text embeddings through cross-modal contrastive learning. Nevertheless, some subtle multi-modal manipulations cause minor semantic inconsistency between two modalities, which are hardly unveiled by normal contrastive learning. To emphasize the semantic inconsistency caused by manipulations, \textbf{HAMMER} proposes manipulation-aware contrastive learning on image and text embeddings. Different from normal cross-modal contrastive learning pulling embeddings of original image-text pairs close while only pushing those of unmatched pairs apart, manipulation-aware contrastive learning pushes away embeddings of manipulated pairs as well so that semantic inconsistency produced by them can be further emphasized. Following InfoNCE loss~\cite{oord2018representation}, we formulate image-to-text contrastive loss by:
\begin{equation}
\footnotesize
\begin{split}
\mathcal{L}_{v2t}(I, T^{+}, T^{-}) = -\mathbb{E}_{p(I,T)}\left[ {\rm log} \frac{{\rm exp}(S(I, T^{+})/\tau)}{\sum\nolimits_{k=1}^{K} {\rm exp} (S(I, T^{-}_{k})/\tau)} \right]
\end{split}
\end{equation}
where $\tau$ is a temperature hyper-parameter, $T^{-} = \{ T^{-}_1, ..., T^{-}_K\}$ is a set of negative text samples that are not matched to $I$ \textit{as well as that belong to manipulated image-text pairs}. Since \texttt{[CLS]} token serves as the semantic representation of the whole image and text, we use two projection heads $h_v$ and $\hat{h}_t$ to map \texttt{[CLS]} tokens of two modalities to a lower-dimensional (256) embedding space for similarity calculation: $S(I, T)=h_v(v_{\rm cls})^{\rm T}\hat{h}_t(\hat{t}_{\rm cls})$. Inspired by MoCo~\cite{he2020momentum}, we learn momentum uni-modal encoders $\hat{E}_v$, $\hat{E}_t$ (an exponential-moving-average version) and momentum projection heads for two modalities respectively. Two queues are used to store the most recent $K$ image-text pair embeddings. Here $\hat{t}_{\rm cls}$ are $\texttt{[CLS]}$ tokens from text momentum encoders and $\hat{h}_t(\hat{t}_{\rm cls})$ means projected text embeddings from text momentum projection head. Similarly, text-to-image contrastive loss is as follows:
\begin{equation}
\footnotesize
\begin{split}
\mathcal{L}_{t2v}(T, I^{+}, I^{-}) = -\mathbb{E}_{p(I,T)}\left[ {\rm log} \frac{{\rm exp}(S(T, I^{+})/\tau)}{\sum\nolimits_{k=1}^{K} {\rm exp} (S(T, I^{-}_{k})/\tau)} \right]
\end{split}
\end{equation}
where $I^{-} = \{ I^{-}_1, ..., I^{-}_K\}$ is a queue of $K$ recent negative image samples that are not matched to $T$ \textit{as well as that belong to manipulated image-text pairs}. $S(T, I)=h_t(t_{\rm cls})^{\rm T}\hat{h}_v(\hat{v}_{\rm cls})$. Inspired by~\cite{yang2022vision}, to maintain reasonable semantic relation within each single modality, we further carry out intra-modal contrastive learning in both modalities. We incorporate all the losses to form Manipulation-Aware Contrastive Loss as follows:
\begin{equation}
\footnotesize
\begin{split}
\mathcal{L}_{MAC} = & \frac{1}{4} [\mathcal{L}_{v2t}(I, T^{+}, T^{-}) + \mathcal{L}_{t2v}(T, I^{+}, I^{-}) + \\
& \mathcal{L}_{v2v}(I, I^{+}, I^{-})+\mathcal{L}_{t2t}(T, T^{+}, T^{-}) ]
\end{split}
\end{equation}

\noindent \textbf{Manipulated Image Bounding Box Grounding.}
As mentioned above, FS or FA swaps identities or edits attributes of faces in images. This alters their correlation to corresponding texts in terms of persons' names or emotions. Given this, we argue that the manipulated image region could be located by finding local patches that have inconsistencies with the text embeddings. In this regard, we perform cross-attention between image and text embeddings to obtain patch embeddings that contain image-text correlation. Attention function~\cite{vaswani2017attention} is performed on normalized query ($Q$), key ($K$), and value ($V$) features as:
\begin{equation}
\vspace*{-2mm}
\footnotesize
\begin{split}
\text{Attention}(Q, K, V) =\text{Softmax}(K^T Q / \sqrt{D})V
\end{split}
\end{equation}
Here we cross-attend the image embedding with text embedding, by treating $Q$ as image embedding, $K$ and $V$ as text embedding as follows:
\begin{equation}
\vspace*{-2mm}
\footnotesize
\begin{split}
U_v(I) & = \text{Attention}(E_v(I), E_t(T), E_t(T)) + E_v(I)
\end{split}
\end{equation}
where $U_v(I)=\{u_{\rm cls}, u_{\rm pat} \}$. $u_{\rm pat} = \{u_1, ..., u_{\rm N} \}$ are $N$ image patch embeddings interacted with text information. Unlike \texttt{[CLS]} token $u_{\rm cls}$, the patch tokens $u_{\rm pat}$ are generated with position encoding~\cite{vaswani2017attention}. This means they possess richer local spatial information and thus are more suitable for manipulated image bbox grounding. Based on this analysis, we propose \textit{Local Patch Attentional Aggregation (LPAA) }to aggregate the spatial information in $u_{\rm pat}$ via an attentional mechanism. This aggregation is performed by cross-attending a $\texttt{[AGG]}$ token with $u_{\rm pat}$ as follows:
\begin{equation}
\vspace*{-2mm}
\footnotesize
\begin{split}
u_{\rm agg} = \text{Attention}(\texttt{[AGG]}, u_{\rm pat}, u_{\rm pat})
\end{split}
\end{equation}
Different from previous work~\cite{zeng2021multi} directly using \texttt{[CLS]} token for bbox prediction, we perform the manipulated bbox grounding based on the attentionally aggregated embedding $u_{\rm agg}$. Specifically, we input $u_{\rm agg}$ into BBox Detector $D_v$ and calculate Image Manipulation Grounding Loss by combing normal $\ell_{1}$ loss and generalized Intersection over Union (IoU) loss~\cite{rezatofighi2019generalized} as follows:
\begin{equation}
\vspace*{-3mm}
\footnotesize
\begin{split}
\mathcal{L}_{IMG} = & \mathbb{E}_{(I,T)\sim P} [\| {\rm Sigmoid}(D_v(u_{\rm agg}))-y_{box}\| \\
& + \mathcal{L}_{\rm IoU}({\rm Sigmoid}(D_v(u_{\rm agg}))-y_{box}) ]
\end{split}
\end{equation}

\begin{table*}[t]
\scriptsize
\centering
\caption{Comparison of multi-modal learning methods for DGM$^{4}$.}
\vspace*{-3mm}
\begin{tabular}{c|ccc|ccc|ccc|ccc}
\toprule[1pt]
Categories & \multicolumn{3}{c|}{Binary Cls}    & \multicolumn{3}{c|}{Multi-Label Cls}  & \multicolumn{3}{c|}{Image Grounding} & \multicolumn{3}{c}{Text Grounding} \\ \hline
Methods              & AUC            & EER            & ACC            & mAP            & CF1            & OF1            & IoUmean       & IoU50        & IoU75        & Precision      & Recall         & F1             \\ \hline
CLIP~\cite{radford2021learning}                 & 83.22          & 24.61          & 76.40          & 66.00          & 59.52          & 62.31          & 49.51           & 50.03          & 38.79          & 58.12          & 22.11          & 32.03          \\
ViLT~\cite{kim2021vilt}                 & 85.16          & 22.88          & 78.38          & 72.37          & 66.14          & 66.00          & 59.32           & 65.18          & 48.10          & 66.48          & 49.88          & 57.00          \\
\rowcolor[HTML]{E3DCDC} 
\textbf{Ours}        & \textbf{93.19} & \textbf{14.10} & \textbf{86.39} & \textbf{86.22} & \textbf{79.37} & \textbf{80.37} & \textbf{76.45}  & \textbf{83.75} & \textbf{76.06} & \textbf{75.01} & \textbf{68.02} & \textbf{71.35} \\ \bottomrule[1pt]
\end{tabular}
\label{tab:multi-modal} 
\vspace*{-4.5mm}
\end{table*}

\subsection{Deep Manipulation Reasoning}
Manipulated token grounding is a tougher task than manipulated bbox grounding as it requires deeper analysis and reasoning on the correlation between images and texts. For example, as illustrated in Fig.~\ref{fig:framework}, we are able to detect the manipulated tokens in $T$, \textit{i.e.,} `force' and `resign', only when we are aware of such negative words mismatching the positive emotion (\textit{i.e.,} smiling faces) in $I$. Besides, we need to summarize multi-modal information to detect fine-grained manipulation types and binary classes. This demands a comprehensive information summary at this stage. To this end, we propose deep manipulation reasoning.

\noindent \textbf{Manipulated Text Token Grounding.}
To model deeper multi-modal interaction, as depicted in Fig.~\ref{fig:framework}, we propose modality-aware cross-attention to further lead text embedding $E_t(T)$ to interact with image embedding $E_v(I)$ through multiple cross-attention layers in Multi-Modal Aggregator $F$. This generates aggregated multi-modal embedding $F(E_v(I), E_t(T))=\{m_{\rm cls}, m_{\rm tok}\}$. In particular, $m_{\rm tok} = \{m_1,..., m_{\rm M} \}$ represent the deeper aggregated embeddings corresponding to each token in $T$. At this stage, each token in $T$ has passed through multiple self-attention layers in $E_t$ and cross-attention layers in $F$. In this way, each token embedding in $m_{\rm tok}$ not only entirely explores the context information of text, but also fully interacts with image features, which fits manipulated text tokens grounding. Moreover, grounding manipulated tokens is equal to labeling each token as real or fake. This is similar to sequence tagging task in NLP. Notably, unlike existing sequence tagging task mainly studied in text modality, manipulated text token grounding here can be regarded as a novel \textit{multi-modal sequence tagging} since each token is interacted with two modality information. In this case, we use a Token Detector $D_t$ to predict the label of each token in $m_{\rm tok}$ and calculate cross-entropy loss as follows:
\begin{equation}
\vspace*{-1mm}
\footnotesize
\begin{split}
\mathcal{L}_{tok} = \mathbb{E}_{(I,T)\sim P}{\rm \textbf{H}}(D_t(m_{\rm tok}), y_{\rm tok})
\end{split}
\end{equation}
where ${\rm \textbf{H}(\cdot)}$ is the cross-entropy function. As mentioned, news on the web is usually noisy with texts unrelated to paired images~\cite{li2021align}. To alleviate over-fitting to noisy texts, as shown in Fig.~\ref{fig:framework}, we further learn momentum versions for Multi-Modal Aggregator and Token Detector, respectively, denoted as $\hat{F}$ and  $\hat{D}_t$. We can obtain the multi-modal embedding from momentum modules as $\hat{F}(\hat{E}_v(I), \hat{E}_t(T))=\{\hat{m}_{\rm cls}, \hat{m}_{\rm tok}\}$. Based on this, momentum Token Detector generates soft pseudo-labels to modulate the original token prediction, by calculating the KL-Divergence as follows:
\begin{equation}
\footnotesize
\begin{split}
\mathcal{L}_{tok}^{mom} = \mathbb{E}_{(I,T)\sim P}{\rm KL} \left[D_t(m_{\rm tok})\| \hat{D}_t(\hat{m}_{\rm tok}) \right]
\end{split}
\end{equation}
The final Text Manipulation Grounding Loss is a weighted combination as follows:
\begin{equation}
\vspace*{-1mm}
\footnotesize
\begin{split}
\mathcal{L}_{TMG} = (1- \alpha)\mathcal{L}_{tok}+\alpha \mathcal{L}_{tok}^{mom}
\end{split}
\end{equation}

\noindent \textbf{Fine-Grained Manipulation Type Detection and Binary Classification.}
Unlike current forgery detection works mainly performing real/fake binary classification, we expect our model to provide more interpretation for manipulation detection. As mentioned in Sec.~\ref{sec:Dataset_Manipulation}, two image and two text manipulation approaches are introduced in \textbf{DGM$^4$} dataset. Given this, we aim to further detect four fine-grained manipulation types. As different manipulation types could appear in one image-text pair simultaneously, we treat this task as a specific \textit{multi-modal multi-label classification}. Since \texttt{[CLS]} token $m_{\rm cls}$ aggregates multi-modal information after modality-aware cross-attention, it can be utilized as a comprehensive summary of manipulation characteristics. We thus concatenate a Multi-Label Classifier $C_m$ on top of it to calculate Multi-Label Classification Loss:
\begin{equation}
\vspace*{-1mm}
\footnotesize
\begin{split}
\mathcal{L}_{MLC} = \mathbb{E}_{(I,T)\sim P}{\rm \textbf{H}}(C_m(m_{\rm cls}), y_{\rm mul})
\end{split}
\end{equation}
Naturally, we also conduct a normal binary classification based on $m_{\rm cls}$ as follows:
\begin{equation}
\vspace*{-3mm}
\footnotesize
\begin{split}
\mathcal{L}_{BIC} = \mathbb{E}_{(I,T)\sim P}{\rm \textbf{H}}(C_b(m_{\rm cls}), y_{\rm bin})
\end{split}
\end{equation}

\section{Experiments}
\label{sec:experiments}

Please refer to appendix for implementation details and rigorous setup of evaluation metrics.

\subsection{Benchmark for DGM$^{4}$}

\begin{table}[t]
\scriptsize
\centering
\caption{Comparison of deepfake detection methods for DGM$^{4}$.}
\vspace*{-4mm}
\begin{tabular}{c|ccc|ccc}
\toprule[1pt]
Categories     & \multicolumn{3}{c|}{Binary Cls}               & \multicolumn{3}{c}{Image Grounding}         \\ \hline
Methods        & AUC            & EER            & ACC            & IoUmean      & IoU50        & IoU75        \\ \hline
TS~\cite{luo2021generalizing} & 91.80          & 17.11          & 82.89          & 72.85          & 79.12          & 74.06          \\
MAT~\cite{zhao2021multi}            & 91.31          & 17.65          & 82.36          & 72.88          & 78.98          & 74.70          \\
\rowcolor[HTML]{E3DCDC} 
\textbf{Ours}  & \textbf{94.40} & \textbf{13.18} & \textbf{86.80} & \textbf{75.69} & \textbf{82.93} & \textbf{75.65}\\ \bottomrule[1pt]
\end{tabular}
\label{tab:deepfake} 
\vspace*{-2mm}
\end{table}

\begin{table}[t]
\scriptsize
\centering
\caption{Comparison of sequence tagging methods for DGM$^{4}$.}
\vspace*{-3mm}
\begin{tabular}{c|ccc|ccc}
\toprule[1pt]
Categories    & \multicolumn{3}{c|}{Binary Cls}               & \multicolumn{3}{c}{Text Grounding}         \\ \hline
Methods       & AUC            & EER            & ACC            & Precision      & Recall         & F1             \\ \hline
BERT~\cite{devlin2019bert}          & 80.82          & 28.02          & 68.98          & 41.39          & 63.85          & 50.23          \\
LUKE~\cite{yamada2020luke}          & 81.39          & 27.88          & 76.18          & 50.52          & 37.93          & 43.33          \\
\rowcolor[HTML]{E3DCDC} 
\textbf{Ours} & \textbf{93.44} & \textbf{13.83} & \textbf{87.39} & \textbf{70.90} & \textbf{73.30} & \textbf{72.08} \\ \bottomrule[1pt]
\end{tabular}
\label{tab:seqtag} 
\vspace*{-2mm}
\end{table}

\begin{table}[t]
\scriptsize
\centering
\caption{Ablation study of image modality.}
\vspace*{-3mm}
\begin{tabular}{c|ccc|ccc}
\toprule[1pt]
Categories    & \multicolumn{3}{c|}{Binary Cls}               & \multicolumn{3}{c}{Image Grounding}              \\ \hline
Methods       & AUC            & EER            & ACC            & IoUmean        & IoU50          & IoU75          \\ \hline
Ours-Image   & 93.96          & 13.83          & 86.13          & 75.58         & 82.44          & \textbf{75.80 }          \\
\textbf{Ours}  & \textbf{94.40} & \textbf{13.18} & \textbf{86.80} & \textbf{75.69} & \textbf{82.93} & 75.65 \\ \bottomrule[1pt]
\end{tabular}
\label{tab:abl_img} 
\vspace*{-2mm}
\end{table}

\begin{table}[t]
\scriptsize
\centering
\caption{Ablation study of text modality.}
\vspace*{-3mm}
\begin{tabular}{c|ccc|ccc}
\toprule[1pt]
Categories    & \multicolumn{3}{c|}{Binary Cls}               & \multicolumn{3}{c}{Text Grounding}         \\ \hline
Methods       & AUC            & EER            & ACC            & Precision      & Recall         & F1             \\ \hline
Ours-Text    & 75.67          & 32.46          & 72.17          & 42.99          & 33.68          & 37.77          \\
\textbf{Ours} & \textbf{93.44} & \textbf{13.83} & \textbf{87.39} & \textbf{70.90} & \textbf{73.30} & \textbf{72.08} \\ \bottomrule[1pt]
\end{tabular}
\label{tab:abl_txt} 
\vspace*{-4mm}
\end{table}

\begin{table*}[t]
\scriptsize
\centering
\caption{Ablation study of losses in the proposed method.}
\vspace*{-3mm}
\begin{tabular}{p{0.45cm}<{\centering}p{0.45cm}<{\centering}p{0.45cm}<{\centering}p{0.45cm}<{\centering}p{0.45cm}<{\centering}|ccc|ccc|ccc|ccc}
\toprule[1pt]
\multicolumn{5}{c|}{Losses}                                                                                                                                   & \multicolumn{3}{c|}{Binary Cls}               & \multicolumn{3}{c|}{Multi-Label Cls}             & \multicolumn{3}{c|}{Image Grounding}            & \multicolumn{3}{c}{Text Grounding}               \\ \hline
BIC                           & MLC                           & MAC                           & IMG                           & TMG                           & AUC            & EER            & ACC            & mAP            & CF1            & OF1            & IoUmean      & IoU50        & IoU75        & Precision      & Recall         & F1             \\ \hline
\CheckmarkBold &    &    &    &    & 91.04          & 16.91          & 83.81          & 20.79          & 33.84         & 33.48          & 4.81          & 0.33           & 0.00           & 15.95          & \textbf{78.70}          & 26.53          \\
\CheckmarkBold &    & \CheckmarkBold & \CheckmarkBold & \CheckmarkBold & 91.74          & 16.08          & 84.39          & 27.22          & 30.81          & 27.62          & 74.05         & 81.34          & 72.59          & 74.30          & 66.84          & 70.37          \\
\CheckmarkBold & \CheckmarkBold &    & \CheckmarkBold & \CheckmarkBold & 92.77          & 14.53          & 86.01          & 85.52          & 79.09          & 79.86          & 75.98         & 83.37          & 75.25          & \textbf{77.82} & 61.83          & 68.91          \\
\CheckmarkBold & \CheckmarkBold & \CheckmarkBold &    & \CheckmarkBold & \textbf{93.21}          & 14.30           & 86.28          & \textbf{86.29} & 79.37 & 80.32 & 4.69          & 0.17           & 0.00         & 75.72          & 67.44 & 71.34 \\
\CheckmarkBold & \CheckmarkBold & \CheckmarkBold & \CheckmarkBold &    & 92.99           & 14.62          & 86.15          & 86.06          & 79.06         & 79.93          & \textbf{76.51} & 83.73 & 76.05          & 13.93          & 58.87          & 22.53           \\
\CheckmarkBold & \CheckmarkBold & \CheckmarkBold & \CheckmarkBold & \CheckmarkBold & 93.19 & \textbf{14.10} & \textbf{86.39} & 86.22          & \textbf{79.37}          & \textbf{80.37}          & 76.45         & \textbf{83.75}          & \textbf{76.06} & 75.01          & 68.02          & \textbf{71.35}          \\ \bottomrule[1pt]
\end{tabular}
\label{tab:abl_loss} 
\end{table*}


\noindent \textbf{Comparison with multi-modal learning methods.} 
We adapt two SOTA multi-modal learning methods to \textbf{DGM$^{4}$} setting for comparison. Specifically, CLIP~\cite{radford2021learning} is one of the most popular \textit{dual-stream} approaches where two modalities are not concatenated at the input level. For adaptation, we make outputs of two streams interact with each other through cross-attention layers. Detection and grounding heads are further integrated on top of them. In addition, ViLT~\cite{kim2021vilt} is a representative \textit{single-stream} approach where cross-modal interaction layers are operated on a concatenation of image and text inputs. We also adapt it by concatenating detection and grounding heads on corresponding outputs of the model. We tabulate comparison results in Table~\ref{tab:multi-modal}. The results show that the proposed method significantly outperforms both baselines in terms of all evaluation metrics. This demonstrates that hierarchical manipulation reasoning is more able to accurately and comprehensively model the correlation between images and texts and capture semantically inconsistency caused by manipulation, contributing to better manipulation detection and grounding.

\noindent \textbf{Comparison with deepfake detection and sequence tagging methods.}
We compare our method with competitive uni-modal methods in two single-modal forgery data splits, respectively. For a fair comparison, in addition to the original ground-truth regarding binary classification, we further integrate manipulation grounding heads into uni-modal models with corresponding annotations of grounding. For image modality, we tabulate the comparison with two SOTA deepfake detection methods in Table~\ref{tab:deepfake}. For text modality, we compare two widely-used sequence tagging methods in NLP to ground manipulated tokens along with binary classification. We report the comparison results in Table~\ref{tab:seqtag}. Tables~\ref{tab:deepfake} and~\ref{tab:seqtag} show that \textbf{HAMMER} performs better than uni-modal methods for single-modal forgery detection by a large margin. This indicates our method trained with multi-modal media also achieves promising manipulation detection and grounding performance in each single modality.

\subsection{Experimental Analysis}
\noindent \textbf{Ablation study of two modalities.} To validate the importance of multi-modal correlation for our model, we perform ablation study by only keeping corresponding input and network components with respect to image (Ours-Image) or text (Ours-Text) modality. We tabulate results in Tables~\ref{tab:abl_img} and~\ref{tab:abl_txt}, showing the performance of complete version of our model surpasses its ablated parts, especially in text modality. This suggests the performance degrades once one of the two modalities is missing without cross-modal interaction. This is to say, through exploiting correlation between two modalities via our model, more complementary information between them can be dug out to promote our task. Particularly, this correlation is more essential for manipulation detection and grounding in text modality.

\noindent \textbf{Ablation study of losses.} 
The considered losses and corresponding results obtained for each case are tabulated in Table~\ref{tab:abl_loss}. As evident from Table~\ref{tab:abl_loss}, removing the task-general loss, \textit{i.e.,} $\mathcal{L}_{MAC}$, nearly all the performance degenerates. This implies manipulation-aware contrastive learning is indispensable for our task. After getting rid of any one of task-specific losses, \textit{i.e.,} $\mathcal{L}_{MLC}$, $\mathcal{L}_{IMG}$ and $\mathcal{L}_{TMG}$, not only the performance of the corresponding task degrades dramatically, but also the overall binary classification performance probably becomes lower. Comparatively, our model with the complete loss function obtains the best performance in most of cases, indicating the effectiveness and complementarity of all losses. In particular, the first row of Table~\ref{tab:abl_loss} represents the current multi-modal misinformation detection scenario where only $\mathcal{L}_{BIC}$ is used. Our method substantially outperforms this baseline on binary classification, implying more manipulation grounding tasks in \textbf{DGM$^{4}$} facilitate binary classification as well.

\begin{figure}[t]
\vspace*{-5mm}
    \centering
     \begin{minipage}[t]{0.55\linewidth}
     \captionsetup{width=0.95\textwidth}
        \centering
        \includegraphics[height=90pt]{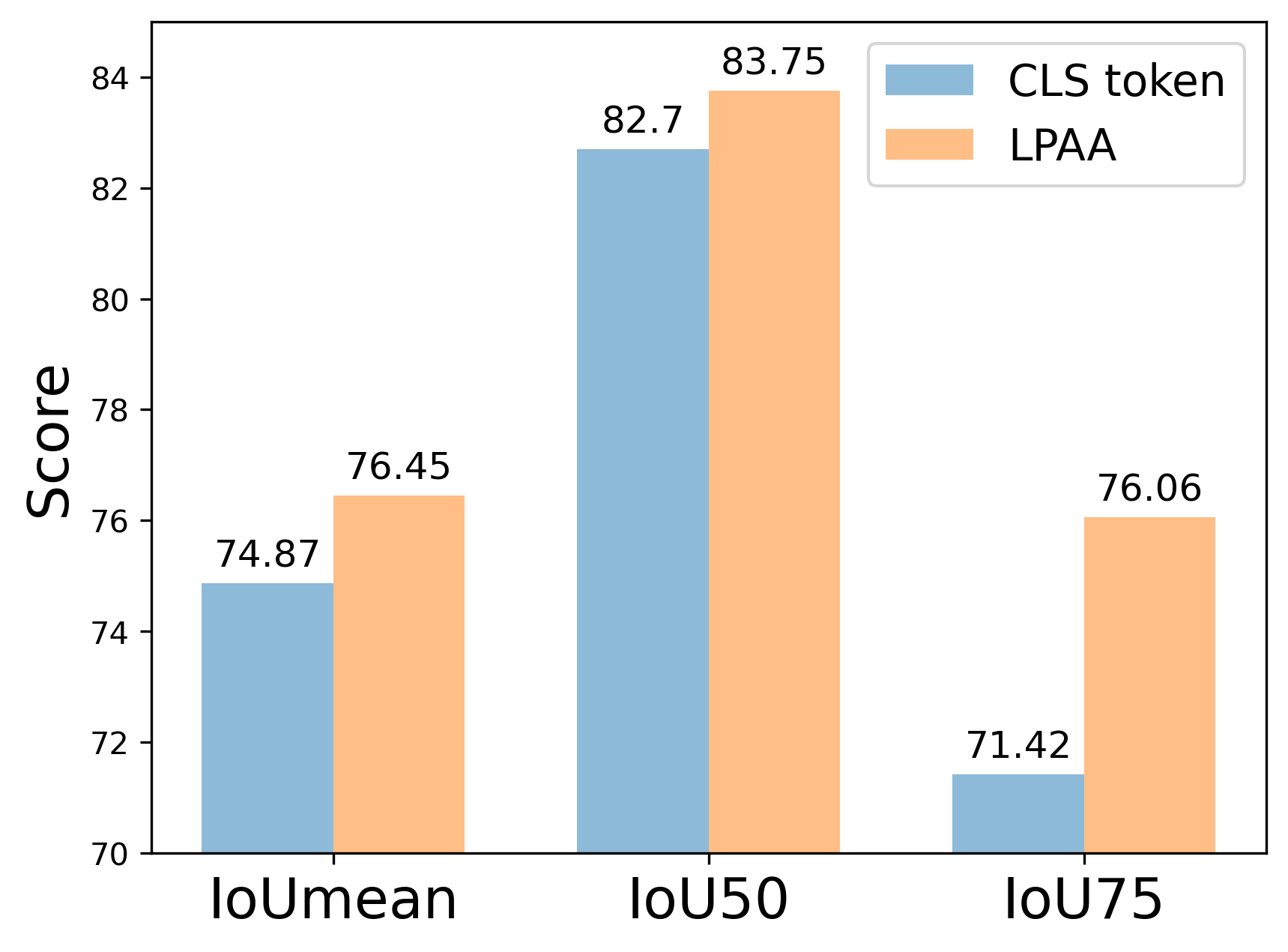}
        \vspace*{-4mm}
        \caption{Efficacy of local patch attentional aggregation (LPAA).}
        \label{fig:abl_lpaa}
    \end{minipage}
    \begin{minipage}[t]{0.40\linewidth}
        \centering
        \includegraphics[height=90pt]{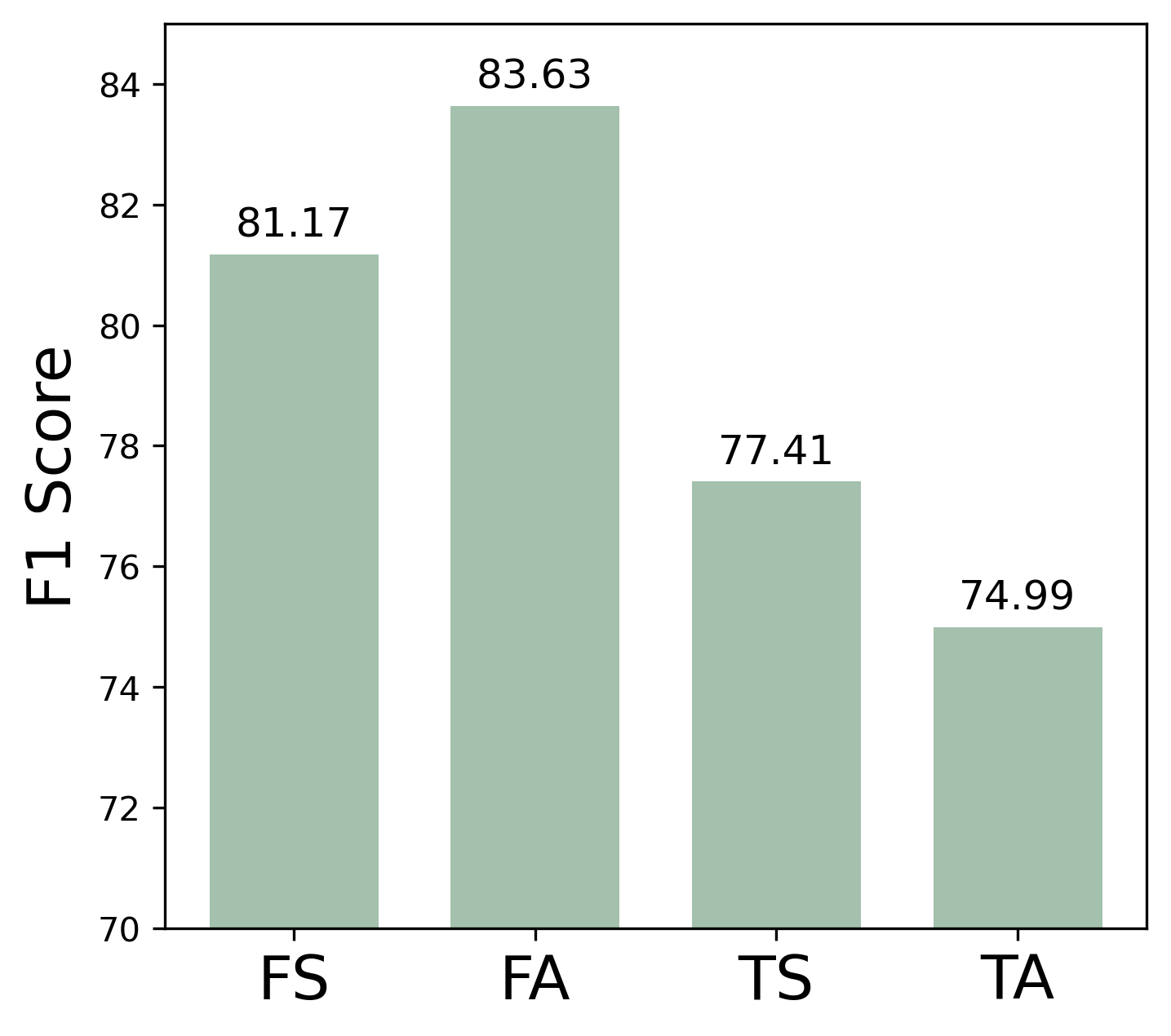}
        \vspace*{-8mm}
        \caption{Performance of each manipulation type.}
        \label{fig:multicls}
    \end{minipage}
    \vspace*{-10mm}
\end{figure}

\noindent \textbf{Efficacy of LPAA.} 
Regarding manipulated bbox grounding, we compare the usage of \texttt{[CLS]} token~\cite{zeng2021multi} with proposed LPAA in Fig.~\ref{fig:abl_lpaa}. Fig.~\ref{fig:abl_lpaa} shows LPAA yields better performance under all metrics, verifying its efficacy.

\noindent \textbf{Details of manipulation type detection.} We plot the classification performance of each manipulation type based on the output of Multi-Label Classifier in Fig.~\ref{fig:multicls}. The results deliver more interpretation that text manipulation detection is harder than image modality and TA is the hardest case.

\begin{figure}[t]
\vspace*{-5mm}
    \centering
     \begin{subfigure}{0.45\linewidth}
        \centering
        \includegraphics[width=1\textwidth]{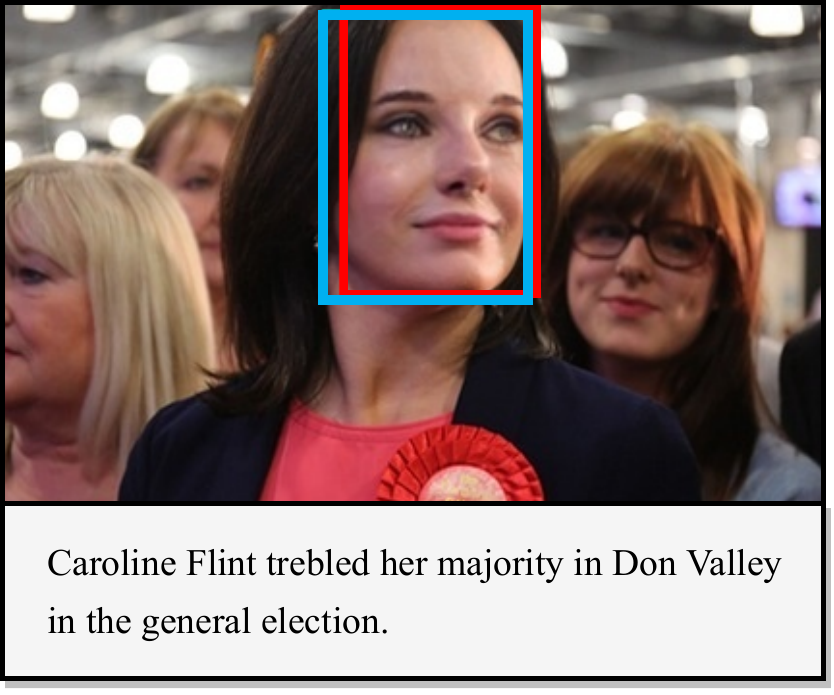}
        \caption{GT: \textcolor{Red}{Fake-FS}, Pred: \textcolor{RoyalBlue}{Fake-FS}}
        \label{fig:vis_FS}
    \end{subfigure}
    \begin{subfigure}{0.45\linewidth}
        \centering
        \includegraphics[width=1\textwidth]{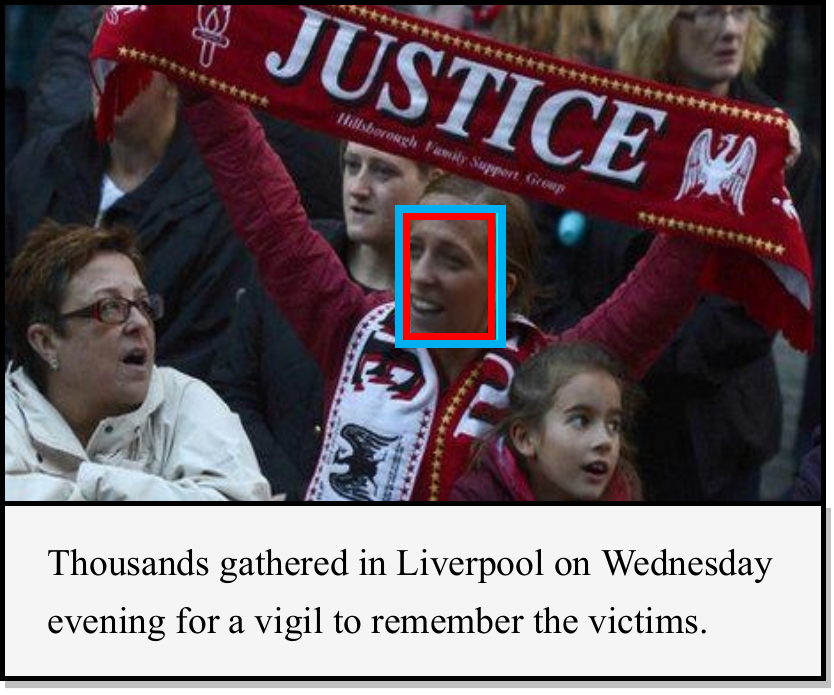}
        \caption{GT: \textcolor{Red}{Fake-FA}, Pred: \textcolor{RoyalBlue}{Fake-FA}}
        \label{fig:vis_FA}
    \end{subfigure}

    \begin{subfigure}{0.45\linewidth}
        \centering
        \includegraphics[width=1\textwidth]{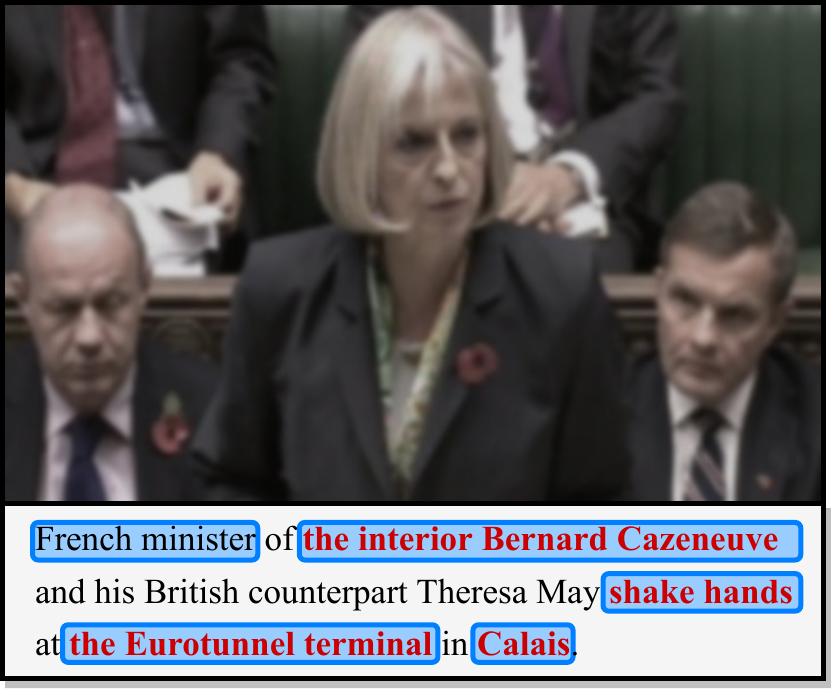}
        \caption{GT: \textcolor{Red}{Fake-TS}, Pred: \textcolor{RoyalBlue}{Fake-TS}}
        \label{fig:vis_TS}
    \end{subfigure}
    \begin{subfigure}{0.45\linewidth}
        \centering
        \includegraphics[width=1\textwidth]{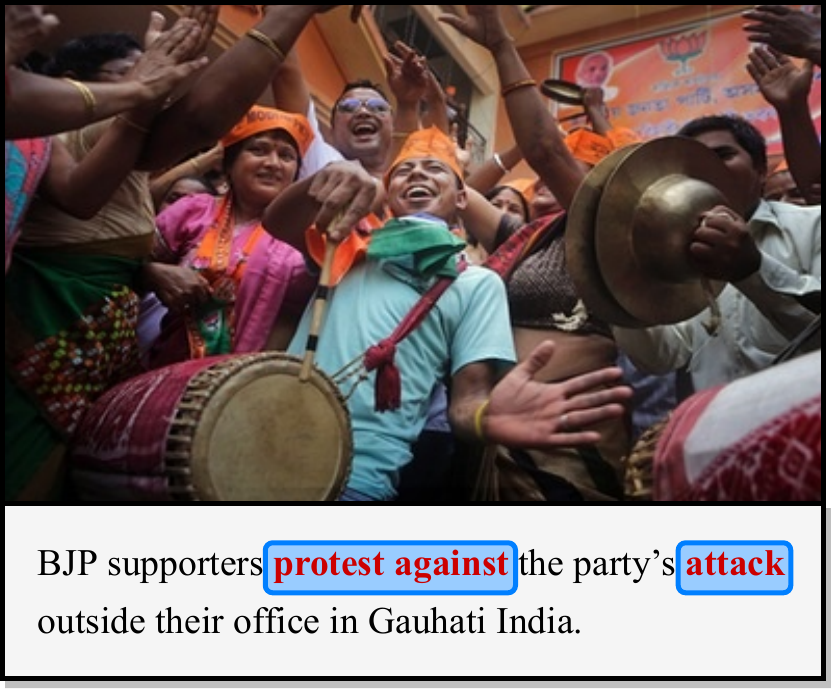}
        \caption{GT: \textcolor{Red}{Fake-TA}, Pred: \textcolor{RoyalBlue}{Fake-TA}}
        \label{fig:vis_TA}
    \end{subfigure}
    \vspace*{-3mm}
    \caption{Visualization of detection and grounding results. Ground truth annotations are in red, and prediction results are in blue.}
    \label{fig:visualization}
\end{figure}

\noindent \textbf{Visualization of manipulation detection and grounding.}
We provide some visualized results of manipulation detection and grounding in Fig.~\ref{fig:visualization}. Fig.~\ref{fig:visualization} (a)-(b) show our method can accurately ground manipulated bboxes and detect correct manipulation types for both FA and FS. Furthermore, most of the manipulated text tokens in TS and all of those in TA are successfully grounded in Fig.~\ref{fig:visualization} (c)-(d). All of them visually verify effective manipulation detection and grounding can be achieved by \textbf{HAMMER}. 

\begin{figure}[t]
\vspace*{-3mm}
    \centering
    \begin{subfigure}{0.45\linewidth}
        \centering
        \includegraphics[width=1\textwidth]{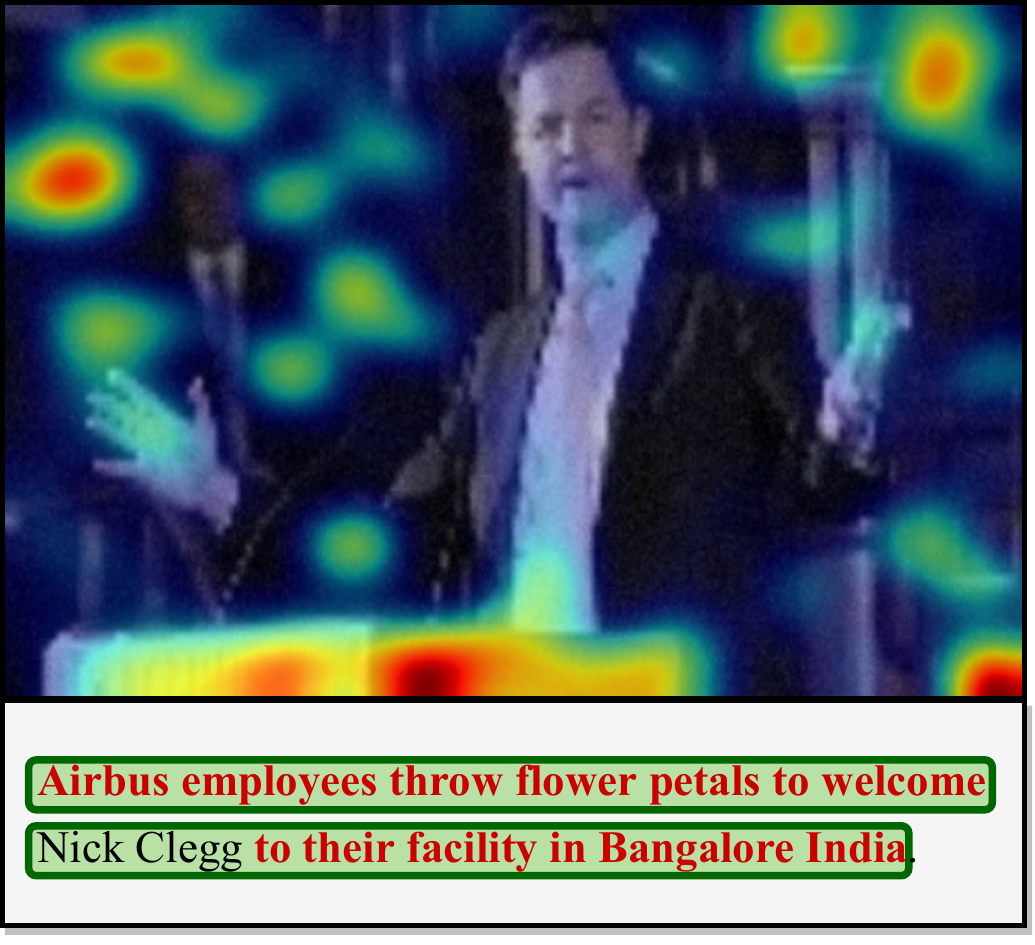}
        \caption{Attention map in TS}
        \label{fig:attn_TS}
    \end{subfigure}
    \begin{subfigure}{0.45\linewidth}
        \centering
        \includegraphics[width=1\textwidth]{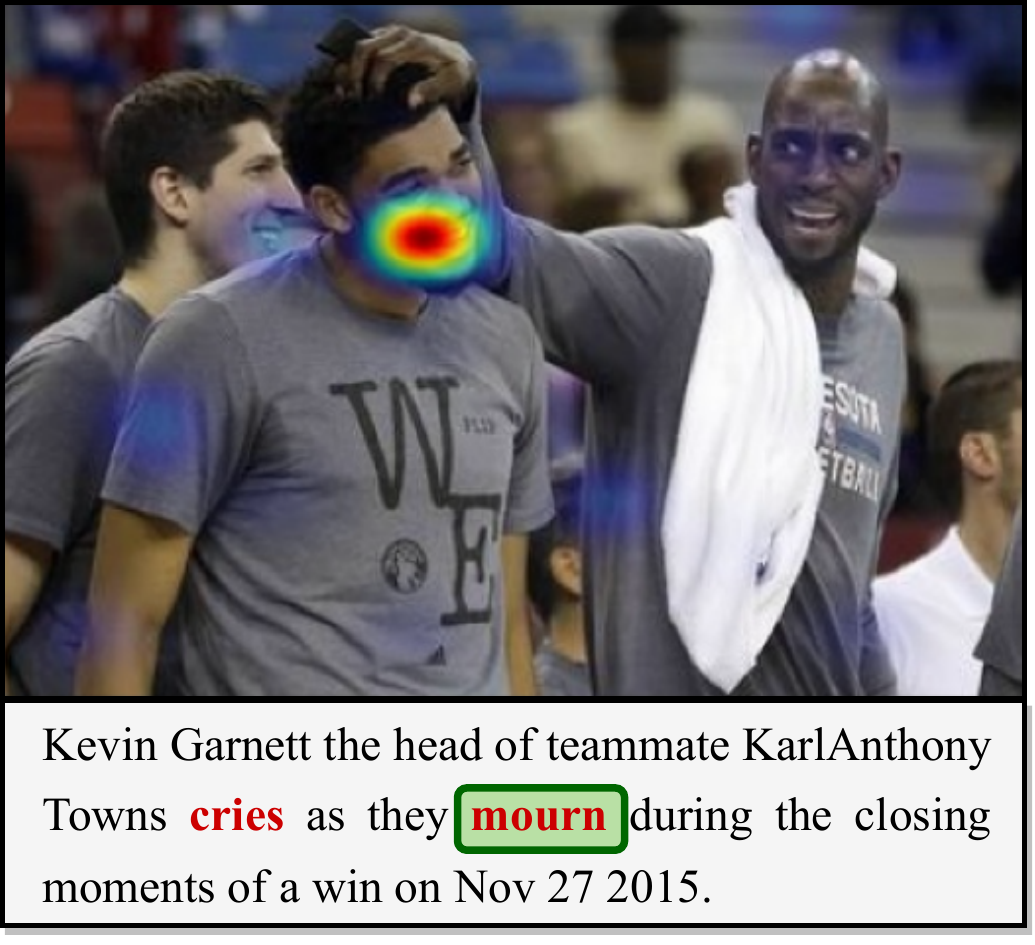}
        \caption{Attention map in TA}
        \label{fig:attn_TA}
    \end{subfigure}
    \vspace*{-3mm}
    \caption{Grad-CAM visualizations on manipulated text tokens.}
    \label{fig:attnmap}
    \vspace*{-4mm}
\end{figure}

\noindent \textbf{Visualization of attention map.}
We provide Grad-CAM visualizations of our model regarding manipulated text tokens in Fig.~\ref{fig:attnmap}. Fig.~\ref{fig:attnmap} (a) shows our model pays attention to surroundings of the character in image. These surroundings indicate the character is giving a speech, which is semantically distinct from text tokens manipulated by TS. As for TA, Fig.~\ref{fig:attnmap} (b) shows the per-word visualization with respect to the manipulated word (`mourn'). It implies our model focuses on the smiling face in image that is semantically inconsistent to the sad sentiment expressed from the manipulated word (`mourn'). These samples prove our model can indeed capture the semantic inconsistency between images and texts to tackle \textbf{DGM$^4$}. 

\section{Conclusion}
This paper studies a novel \textbf{DGM$^4$} problem, aiming to detect and ground multi-modal manipulations. We construct the first large-scale \textbf{DGM$^4$} dataset with rich annotations. A powerful model \textbf{HAMMER} is proposed and extensive experiments are performed to demonstrate its effectiveness.

\section*{Acknowledgements}
This study is supported by the Ministry of Education, Singapore, under its MOE AcRF Tier 2 (MOE-T2EP20221-0012), NTU NAP, and under the RIE2020 Industry Alignment Fund – Industry Collaboration Projects (IAF-ICP) Funding Initiative, as well as cash and in-kind contribution from the industry partner(s).

\clearpage
{\small
\bibliographystyle{ieee_fullname}
\bibliography{egbib}
}

\clearpage
\section*{Supplementary Material}
\appendix

\section{Implementation Details.} 
All of our experiments are performed on 8 NVIDIA V100 GPUs with PyTorch framework~\cite{paszke2017automatic}. Image Encoder is implemented by ViT-B/16~\cite{dosovitskiy2020image} with 12 layers. Both Text Encoder and Multi-Modal Aggregator are built based on a 6-layer transformer initialized by the first 6 layers and the last 6 layers of BERT$_{base}$~\cite{devlin2019bert}, respectively. Binary Classifier, Multi-Label Classifier, BBox Detector, and Token Detector are set up to two Multi-Layer Perception (MLP) layers with output dimensions as 2, 4, 4, and 2. We set the queue size $K = 65, 536$. AdamW~\cite{loshchilov2017decoupled} optimizer is adopted with a weight decay of 0.02. The learning rate is warmed-up to $1e^{-4}$ in the first 1000 steps, and decayed to $1e^{-5}$ following a cosine schedule. 

\section{Evaluation Metrics.}
To evaluate the proposed new research problem DGM$^{4}$ comprehensively, we set up rigorous evaluation protocols and metrics for all the manipulation detection and grounding tasks.
\begin{itemize}[leftmargin=*]
\item \textbf{Binary classification:} Following current deepfake methods~\cite{luo2021generalizing,zhao2021multi}, we adopt Accuracy (ACC), Area Under the Receiver Operating Characteristic Curve (AUC), and Equal Error Rate (EER) for evaluation of binary classification.
\item \textbf{Multi-label classification:} Like existing multi-label classification methods~\cite{ben2020asymmetric,durand2019learning}, we use mean Average Precision (MAP), average per-class F1 (CF1), and average overall F1 (OF1) for evaluating the detection of fine-grained manipulation types.
\item \textbf{Manipulated image bounding box grounding:} To examine the performance of predicted manipulated bbox, we calculate the mean of Intersection over Union (IoUmean) between ground-truth and predicted coordinates of all testing samples. Moreover, we set two thresholds (0.5, 0.75) of IoU and calculate the average accuracy (correct grounding if IoU is above the threshold and versa vice), which are denoted as IoU50 and IoU75.
\item \textbf{Manipulated text token grounding:} Considering the class imbalance scenario that manipulated tokens are much fewer than original tokens, we adopt Precision, Recall, F1 Score as metrics. This contributes to a more fair and reasonable evaluation for manipulated text token grounding.

\end{itemize}

\end{document}